\documentclass{article}
\usepackage{ijcai23}

\usepackage{times}
\usepackage{soul}
\usepackage{url}
\usepackage[hidelinks]{hyperref}
\usepackage[utf8]{inputenc}
\usepackage[small]{caption}
\usepackage{graphicx}
\usepackage{amsmath}
\usepackage{amsthm}
\usepackage{booktabs}
\usepackage{algorithm}
\usepackage{algorithmic}
\usepackage[switch]{lineno}
\usepackage{amssymb}
\usepackage{mathrsfs}
\usepackage{multirow}
\usepackage{graphbox}
\usepackage[misc]{ifsym}

\usepackage{subcaption}

\title{Learning Gaussian Mixture Representations for Tensor Time Series Forecasting
}
\author{
Jiewen Deng$^1$\and
Jinliang Deng$^{1,2}$\and
Renhe Jiang$^{3,1}$* \And
Xuan Song$^{1,3}$
\affiliations
$^1$Southern University of Science and Technology\\
$^2$University of Technology Sydney\\
$^3$The University of Tokyo
\emails
dengjw1@outlook.com,
jinliang.deng@student.uts.edu.au \\
jiangrh@csis.u-tokyo.ac.jp (*corresponding),
songx@sustech.edu.cn
}

\begin{document}

\maketitle

\begin{abstract}
Tensor time series (TTS) data, a generalization of one-dimensional time series on a high-dimensional space, is ubiquitous in real-world scenarios, especially in monitoring systems involving multi-source spatio-temporal data (e.g., transportation demands and air pollutants). Compared to modeling time series or multivariate time series, which has received much attention and achieved tremendous progress in recent years, tensor time series has been paid less effort. Properly coping with the tensor time series is a much more challenging task, due to its high-dimensional and complex inner structure. In this paper, we develop a novel TTS forecasting framework, which seeks to individually model each heterogeneity component implied in the time, the location, and the source variables. We name this framework as \textbf{GMRL}, short for \underline{G}aussian \underline{M}ixture \underline{R}epresentation \underline{L}earning. Experiment results on two real-world TTS datasets verify the superiority of our approach compared with the state-of-the-art baselines. Code and data are published on \url{https://github.com/beginner-sketch/GMRL}.
\end{abstract}

\section{Introduction}

Tensor time series (TTS) data occur as sequences of multidimensional arrays \cite{rogers2013multilinear,jing2021network}. For example, urban public transportation management involves data coming from multiple times, locations, and transportation modes, i.e., Taxi/Bike Inflow/Outflow \cite{zhang2016dnn,zhang2017deep}, based on which compelling predictions are crucial to subsequent decision-making, especially for those urban-sensitive cases. 
\begin{figure}[t]
  \centering
  \includegraphics[width=\linewidth]{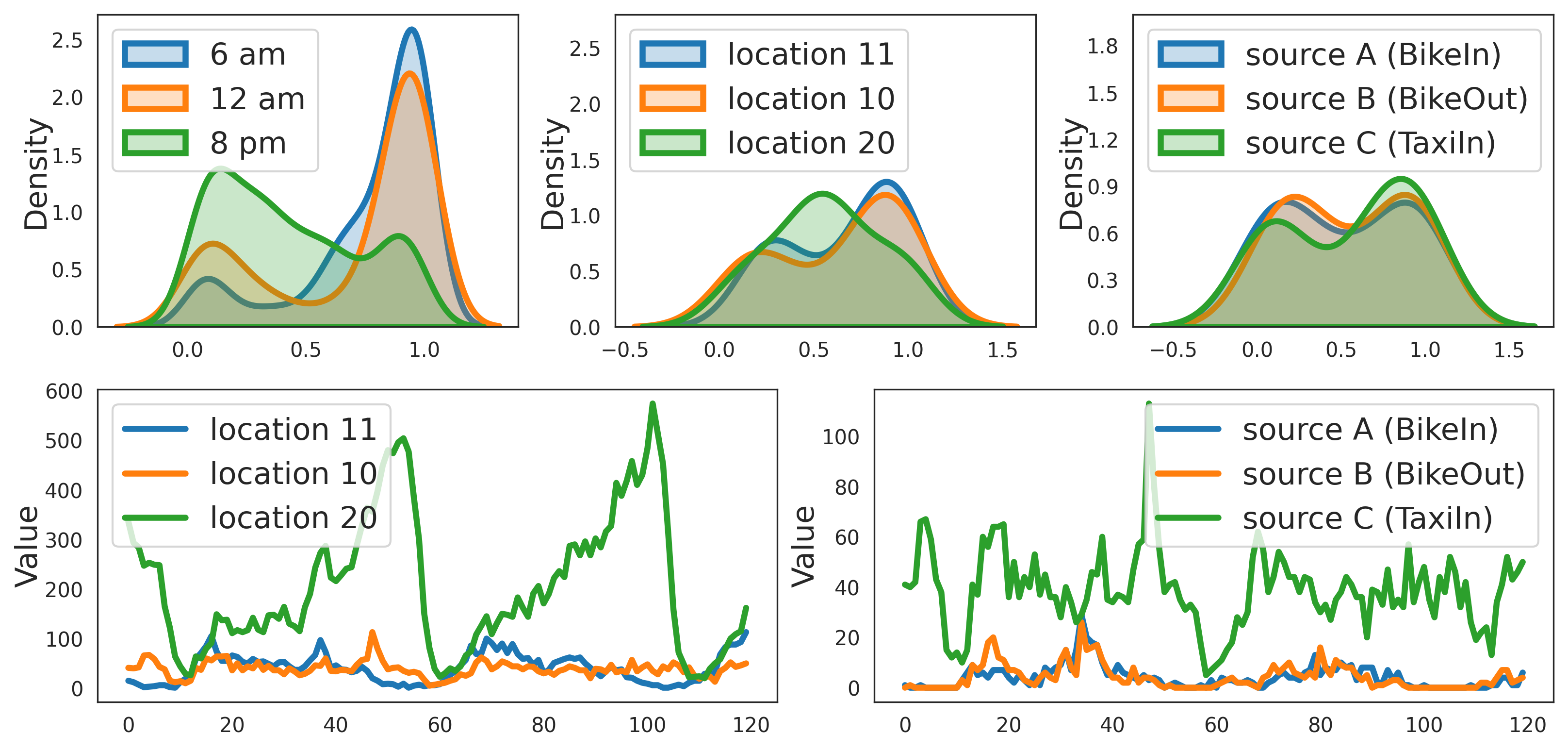}
  \caption{Demonstration of heterogeneous NYC traffic demand data. It contains four sources: Bike Inflow, Bike Outflow, Taxi Inflow, and Taxi Outflow. Regarding probability density (upper), dynamic heterogeneity exists in the time/location/source variables, consistent with the evolution trend (lower).}
  \label{fig:intro}
\end{figure}
Compared with multivariate time series (MTS), TTS involves an additional variable apart from the initial time and location variables, namely the $\emph{source}$ variable. The most significant difference between MTS and TTS is that MTS involves data in a matrix, where the rows represent the time variable and the columns define the location variable. In contrast, TTS concerns tensor data, which usually takes on more substantial heterogeneity and dynamics. Although identifying and extracting similar patterns is reasonable and typical for time series prediction, the components over variables can be dynamic in many real-world applications. Failure to explicitly obtain the quantitative form of patterns may mistakenly cause the model to capture the pattern change in TTS. We use an example of NYC traffic demand data in Figure~\ref{fig:intro} to illustrate it. We collect the historical data of the three frames/locations/sources and plot their data distribution in the upper part of Figure~\ref{fig:intro}. Based on the characteristics (the mean and variance) of the data distributions, we can trivially draw a conclusion that the instance at 6 am is more similar to 12 am than 8 pm. This is consistent with the evolving patterns of observation in the lower part of Figure~\ref{fig:intro}. Such distinguishability over the \emph{time} variable also holds water over the \emph{location} and \emph{source} variables. Given the complex patterns, capturing heterogeneities arising from multiple variables is an essential problem in TTS forecasting.

Deep forecasting models have achieved great success in more effective time series forecasting. Graph neural network (GNN) \cite{wu2020connecting,shao2022decoupled}, temporal convolution networks (TCN) \cite{wu2020connecting}, recurrent neural networks (RNN) \cite{lai2018modeling}, and attention mechanisms \cite{wu2021autoformer,zheng2020gman} are well-used for prediction problems. Each individual or a combination of these manipulations performs well in modeling the characteristics of a single domain or spatial-temporal regularities. Such methods are more prone to focus on low-level hierarchical information and are blind to the rich information and challenges brought by source variables. An increasing number of recent works are drawing forth the evolution from other modes as an auxiliary input, adding ancillary information to borrow similar historical patterns \cite{wu2020hierarchically,fang2021mdtp,han2021dynamic}. These practices, to a certain degree, bring mode-wise awareness. Nevertheless, they mainly help with the spatial-temporal parts, and the heterogeneity or the interactions brought by variables (i.e., time, location, and source) are still underexplored.
Recently, cluster-wise learning breaks the independence between instances by exploiting the latent cluster information among samples. The learned representations are expected to retain higher-level heterogeneous information by taking advantage of additional prior information brought by clustering. However, existing approaches \cite{zhu2021mixseq,duan2022plae}  (i) neglect the dynamic nature of time series, assuming that both the representations and the members of the clusters will remain the same over time and (ii) the interactions over multiple variables as well, which are not applicable for intricate and complex TTS. Thus far, while evolution regularities in time series have been studied intensively, dynamic heterogeneity and interactions of multiple variables in TTS have yet to be adequately tackled.

Therefore, we are motivated to propose a novel TTS learning framework for accurate forecasting, namely \underline{G}aussian \underline{M}ixture \underline{R}epresentation \underline{L}earning (\textbf{GMRL}). It consists of three steps: (i) incorporating the multiple variables information into the model through a TTS embedding (\emph{TTSE}); (ii) designing a dynamic Gaussian mixture representation extractor  to explicitly represent the complicated heterogeneous components among multiple variables; (iii) leveraging techniques of memory that enhances the learned representation to learn to distinguish and generalize to diverse scenarios. The contributions of this work are summarized as follows:
\begin{itemize}
    \item We design a Gaussian Mixture Representation Extractor (GMRE), to explicitly disentangle and quantify the heterogeneity components in the time, the location, and the source variables.
    \item We propose a Hidden Representation Augmenter (HRA) to equip the learned representations with sequence-specific global patterns on the fly, intrinsically improving model adaptability to TTS.
    \item We conduct thorough experiments on two real-world datasets, one is traffic dataset containing four sources of transportation demands (i.e., Taxi/Bike Inflow/Outflow), another is air quality dataset containing three sources of pollutants (i.e., PM2.5, PM10, SO$_2$). The results verify the superiority of our method over the state-of-the-arts.
\end{itemize}

\section{Preliminaries}
In this section, we formulate the tensor time series forecasting problem, and introduce the used notations (in Table~\ref{tab:notations}).

\begin{table}[t]
    \scriptsize
    \centering
    \setlength{\tabcolsep}{1mm}{
    \begin{tabular}{l|l}
        \hline
        Notation  & Description\\
        \hline
        $L$ / $S$ / $T$  & Number of locations/sources/time steps.\\
        $O$ & Number of horizons.\\
        $K$ & Number of cluster components.\\
        $X \in \mathbb{R}^{T \times L\times S}$ / $\hat{Y} \in \mathbb{R}^{O \times L\times S}$ & Input/Output tensor.\\
        \hline
        $E \in \mathbb{R}^{T \times L \times S \times d_z}$ & Tensor time series embedding.\\
        $H \in \mathbb{R}^{T \times L \times S \times d_k}$ & Initial representation. \\
        $\hat{H} \in \mathbb{R}^{T \times L \times S \times d_k}$ & Representation from GMRE. \\
        $H^\text{(gm)} \in \mathbb{R}^{T \times L \times S \times 2d_k}$ & 
        Gaussian mixture representation. \\
        $H^\text{(te)} \in \mathbb{R}^{T \times L \times S \times d_k}$ & Representation from TE. \\
        $H^\text{(sc)} \in \mathbb{R}^{L \times S \times d_k}$ & Representation after skip connection.\\
        $H^\text{(aug)} \in \mathbb{R}^{L \times S \times (d_k+d_m)}$ & Representation augmented by HRA. \\
        \hline
    \end{tabular}}
    \caption{Notations and Descriptions}
    \label{tab:notations}
\end{table}

\noindent \textbf{Problem 1} (Tensor Time Series Forecasting). 
Given $T$ consecutive time steps, $L$ discretized locations, and $S$ sources, the multi-source spatio-temporal data can be aggregated into a \emph{Tensor Time Series} $X \in \mathbb{R}^{T\times L\times S}$. Given $X$, the tensor time series forecasting aims to forecast the next $O$ steps as $\hat{Y} \in \mathbb{R}^{O \times L\times S}$, which can be expressed by the following conditional distribution:
\begin{equation}
P(\hat{Y}|X) = \prod_{t=1}^{O} P(\hat{Y}_{t,:,:}|X)
\end{equation}

\begin{figure*}[t]
  \centering
  \includegraphics[width=\linewidth]{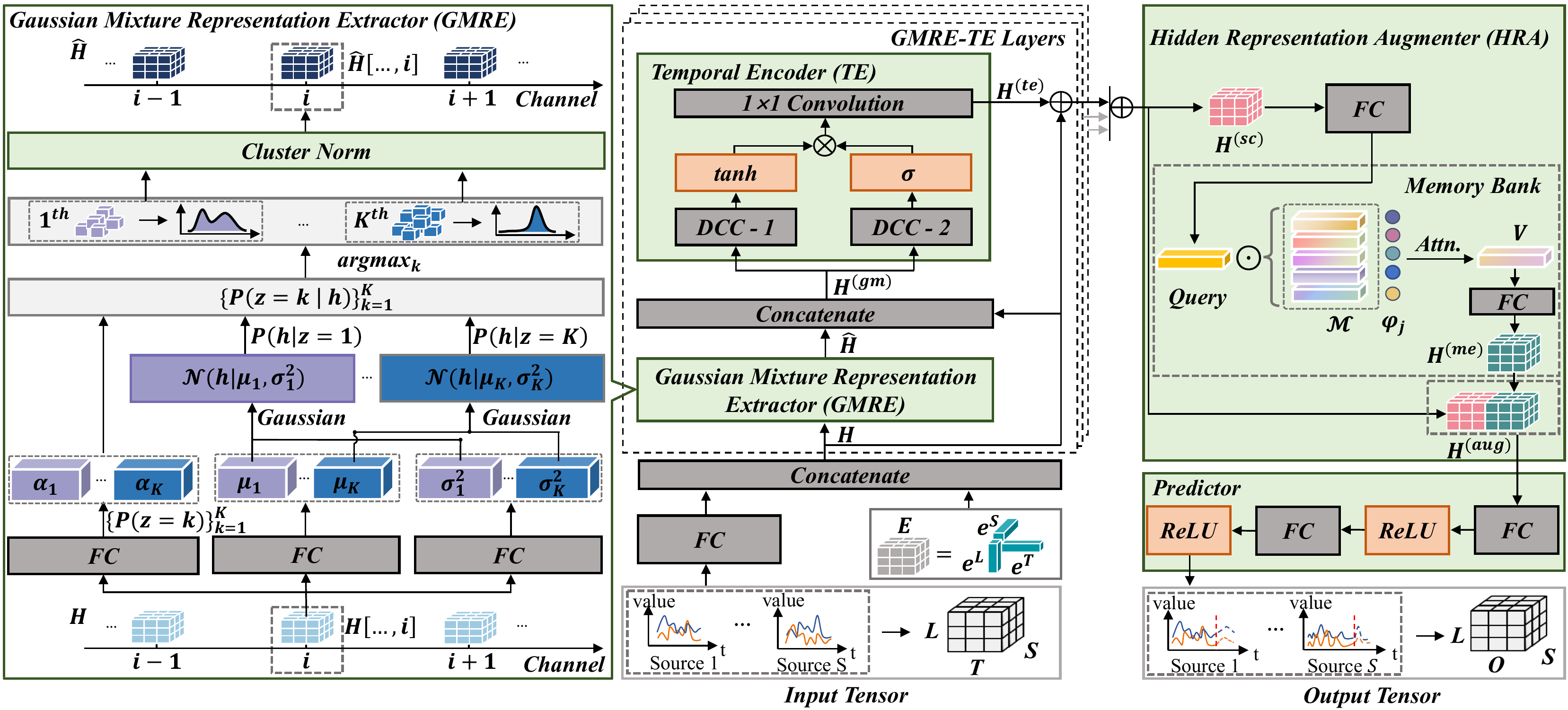}
  \caption{\underline{G}aussian \underline{M}ixture \underline{R}epresentation \underline{L}earning (GMRL): (i) taking a three-dimension TTS as input; (ii) multiple proposed Gaussian Mixture Representation Extractors (GMRE) and Temporal Encoders (TE) are stacked layer by layer; (iii) applying the skip connection to fusion the representation output from each GMRE-TE layer; (iv) a Hidden Representation Augmenter (HRA) is deployed to strengthen representation with global patterns; (v) a Predictor is used to generate the prediction for TTS.}
  \label{fig:framework}
\end{figure*}

\section{Methodology}
In this section, we elaborate on the technical details of the proposed framework \underline{G}aussian \underline{M}ixture \underline{R}epresentation \underline{L}earning (GMRL), as demonstrated in Figure~\ref{fig:framework}.

\subsection{Tensor Time Series Embedding}
Since the evolution patterns of TTS are constrained by the location and the source variables, it is crucial to incorporate the variables' information into the predictive model. Specifically, we generate \emph{location embedding} and \emph{source embedding} to encode corresponding variables into vectors that preserve static information, respectively expressed as $e^{L}_l \in \mathbb{R}^{d_z}$ and $e^{S}_s \in \mathbb{R}^{d_z}$. The above embedding vectors only provide static representations, which could not represent the dynamic correlations among locations and sources. Thus, we further utilize \emph{temporal embedding} $e^{T}_t \in \mathbb{R}^{d_z}$ to encode each historical time step into a vector. To obtain the variable representations, we fuse the aforementioned embedding vectors in an additive manner. For location $l$ and source $s$ at time step $t$, the embedding is defined as: $e_{t,l,s}=e^{T}_t+e^{L}_l+e^{S}_s$. Then we get the TTSE ($E \in \mathbb{R}^{T \times L \times S \times d_z}$) of $L$ locations and $S$ sources in $T$ time steps. Next, the initial representation fed into subsequent modules is $H=[f(X),E]$, where $f(\cdot)$ represents linear transformation and $[,]$ denotes concatenation operation.

\subsection{Gaussian Mixture Representation Extractor}
The canonical cluster analysis method is used to reveal the underlying nature and patterns of data, enabling the classification of the internal correlation structure of data. However, existing time series modeling methods rely on cluster distinctions that do not necessarily indicate an optimal solution. Additionally, time series data feature a dynamic context, implying that the characteristics of the clusters, which describe real-time time series data, will change over time as a function of the dynamic context.

To overcome the above limitations, we develop a dynamic Gaussian Mixture Representation Extractor (GMRE) to explicitly capture the patterns of time series data based on the underlying context. Moreover, we expect different channels to hold information about distinct contextual attributes, i.e., the factors of variation, exhibiting mutual independence and complementarity. For instance, some channels correspond to spatial identities, remaining invariant over time, while others correspond to temporal identities, displaying variations, potentially with different frequencies, over time. Motivated by this, we perform channel-wise clustering instead of the conventional sample-wise clustering, in order to separate the clusters for each channel. A practical advantage of injecting inter-channel independence lies in reducing the number of learnable parameters involved in the model, as modeling interactions among the channels is no longer necessary.

Given a representation $H \in \mathbb{R}^{T \times L \times S \times d_k}$ derived from an input tensor $X$, GMRE aims to learn a mixture of Gaussian distributions over each scalar value within $H$. Technically, the probability distribution is conditioned on the observation (i.e., input tensor). However, for simplicity, we omit this condition in the following probability formulas. Firstly, we introduce a latent variable $z[i] \in [1, K]$ to identify the membership (belonging to which of the $K$ clusters) of each scalar value for each channel $i \in [1, d_k]$. Given $h \in \mathbb{R}^{d_k}$ (one of the $T \times L \times S$ small cubes in Figure~\ref{fig:framework}) from the representation $H$, GMRE infers the probability distribution for each scalar value in $h[i]$ (i.e., $i$-th channel of $h$). Formally, the probability of $h[i]$ can be decomposed as follows:
\begin{equation}
P(h[i])=\sum_{k=1}^K P(z[i]=k) P(h[i]|z[i]=k)
\end{equation}
Then, we can specify the expressions for $P(z[i]=k)$ and $P(h[i]|z[i]=k)$. Formally, we obtain the prior distribution of $z$ as follows:
\begin{align}
P(z[i]=k) &= \alpha_k(H[..., i])\nonumber\\
&= \frac{\exp(W^{\alpha}_{k, i} \;\text{vec}(H[..., i]))}{\sum_{j=1}^K \exp(W^{\alpha}_{j, i} \;\text{vec}(H[..., i]))}
\end{align}
where the membership score associated with the $i^\text{th}$ channel belonging to the $k^\text{th}$ cluster is parameterized by a linear mapping with $W^{\alpha}_{k, i} \in \mathbb{R}^{T L S}$. For $P(h[i]|z[i]=k)$, we assume it follows a Gaussian distribution $\mathcal{N}(h[i]|\mu_{k, i},\sigma_{k, i}^2)$, where the mean $\mu_{k, i} \in \mathbb{R}$ and variance $\sigma_{k, i}^2 \in \mathbb{R}$ are generated by:
\begin{equation}
\left\{
\begin{aligned}
\mu_{k, i} &= W^{\mu}_{k, i}\text{vec}(H[..., i]) +b^{\mu}_{k, i},\\
\sigma^2_{k, i} &= \exp\left(W^{\sigma}_{k, i} \text{vec}\left(H[..., i]\right) +b^{\sigma}_{k, i}\right),
\end{aligned}
\right.
\end{equation}
where $W_{k, i}^{\mu}, W^{\sigma}_{k, i} \in \mathbb{R}^{ T L S}$ are weights to project representation $H[..., i]$ to $\mu_{k, i}$ and $\sigma_{k, i}^2$, and $b^{\mu}_{k, i}, b^{\sigma}_{k, i} \in \mathbb{R}$ are the corresponding biases. $\exp(\cdot)$ ensures that $\sigma^2_{k, i}$ remains greater than zero.

With $P(z[i]=k)$ and $P(h|z[i]=k)$ available, we can derive the posterior probability $P(z[i]=k|h[i])$ as follows, using Bayes' theorem:
\begin{align}
P(z[i]=k|h[i]) &= \frac{P(z[i]=k)P(h[i]|z[i]=k)}{P(h[i])}, \nonumber\\ 
&= \frac{\alpha_k(H[..., i]) \cdot \mathcal{N}(h[i]|\mu_{k, i},\sigma_{k, i}^2)}{\sum_j \alpha_j(H[..., i]) \cdot \mathcal{N}(h[i]|\mu_{j, i},\sigma_{j, i}^2)}
\label{eq:cluster}
\end{align}
Following Eq.~\ref{eq:cluster}, we can assign the $i^\text{th}$ channel of $h$ to cluster $k=\underset{k}{\arg \max} {P(z[i]=k|h[i])}$.

\paragraph{Cluster Normalization.} Inspired by the simplex and efficiency of normalization technology, we devise \emph{Cluster Norm} to increase the distance between clusters while reducing the distance between samples within a cluster. Cluster-wise normalization is applied to each sample belonging to the $k^{\text{th}}$ cluster to get a set of representations as:
\begin{align*}
\hat{H}=&\left[ \frac{{H[t,l,s, i]} - \mu_{k, i}}{\sigma_{k, i} + \varepsilon}
 \right. \\
    & | \left.k=\underset{k}{\arg \max} {P(z[i]=k|H[t,l,s, i])}
\right]_{t, l, s, i}
\end{align*}
\noindent where $\varepsilon$ is a small constant to preserve numerical stability. Then, we can obtain the \underline{G}aussian \underline{m}ixture representation $H^{(\text{gm})}=[H,\hat{H}] \in \mathbb{R}^{T \times L \times S \times 2d_k}$.

\paragraph{Probability Regularization.} To enable $P(h[i])$ to describe the empirical data properly, we must regularize the probability of empirical data under $P(h[i])$. Following the convention, we alternatively minimize minus $\log P(h[i])$, equivalent to maximizing the $P(h[i])$.
\begin{align}
\log P(h[i])
=&\log \sum_{z} P(h[i], z[i])\nonumber\\
=& \sum_{z} P(z[i]|h[i]) \log \frac{P(h[i], z[i])}{P(z[i]|h[i])}\nonumber\\
=&\mathbb{E}_{P(z[i]|h[i])}\log P(h[i]|z[i]) \nonumber\\
&-\text{KL}(P(z[i]|h[i])||P(z[i]))
\label{eq:lb}
\end{align}
\noindent where $KL(\cdot)$ denotes Kullback–Leibler divergence. However, directly maximizing the LB (in Eq.~\ref{eq:lb}) may cause mode collapsing problem, which means the model cannot distinguish features of each cluster in the expected way \cite{zhu2021mixseq}. We import the mutual information $I(z[i],h[i])$ into LB \cite{zhao2018unsupervised} to address this issue. Formally, we derive the following cluster objective to minimize:
\begin{align}
\mathcal{L}^{(\text{cluster})} =& \frac{1}{d_k} \sum_{i} \big[\text{KL}(Q(z[i])||P(z[i])) \nonumber\\
 &-\mathbb{E}_{P_\text{data}(h[i])}\mathbb{E}_{P(z[i]|h[i])}\log P(h[i] | z[i])\big]
\label{eq:cluloss}
\end{align}
where $Q(z[i])=\mathbb{E}_{P_\text{data}(h[i])} P(z[i]|h[i])$ is the expectation of posteriors with respect to the empirical data distribution.

\subsection{Temporal Encoder}
We implement a Temporal Encoder by applying a set of standard extended 1D convolution filters to extract high-level temporal features. 
Specifically, our Temporal Encoder consists of three parts: (i) two Dilated Causal Convolutions (DCC) \cite{Oord2016WaveNetAG} for achieving a larger receptive field for the representation $H^{(\text{gm})}$, i.e., the receptive field size grows in a linear progression with the depth of the network and the kernel size of the filter; (ii) a Gated Linear Unit (GLU) for introducing the nonlinearity into the network; (iii) a $1 \times 1$ convolution function for capturing sequential patterns of time series data. Formally, it works as follows:
\begin{equation}
\resizebox{.91\linewidth}{!}{$
\displaystyle
H^{(\text{te})} = f_{\text{conv}}(\text{tanh}(H^{(\text{gm})} \ast W_{\text{dc1}}) \odot \sigma (H^{(\text{gm})} \ast W_{\text{dc2}}))
$}
\end{equation}
\noindent where $\ast$ denotes the dilated causal convolution, $W_{\text{dc1}}$ and $W_{\text{dc2}}$ are learnable convolution filters for DCC-1 (filter) and -2 (gate), and $\odot$ denotes an element-wise multiplication operator. $f_{\text{conv}}(\cdot)$ denotes a $1 \times 1$ convolution function.

\subsection{Hidden Representation Augmenter}
Here, we design a \emph{Hidden Representation Augmenter} (HRA) to further enhance the learned representation to be aware of and adaptive to various scenarios. HRA is motivated by a line of research on memory networks (\cite{wang2022event,jiang2022spatio}), which has shown promising results in capturing global patterns.
The basic idea of the memory module is to learn a parameterized memory bank that caches global patterns and projects each variable to the same feature space with the memory. The memory module brings two advantages: (i) learning and storing meaningful patterns from a global perspective; (ii) utilizing the knowledge of patterns to construct global representations.

We thereby utilize the idea of a memory module and further put the final representations conditioned on a plugin memory bank $\mathcal{M}$ to encourage the discovery of high-level evolution prototypes, which are representations incorporating the knowledge of the time, the location, and the source variables. Specifically, the memory blank $\mathcal{M} \in \mathbb{R}^{m \times d_m}$ starts by taking the \underline{s}kip \underline{c}onnection representation $H^{(\text{sc})} \in \mathbb{R}^{L \times S \times d_k}$ as a query, where $m$ and $d_m$ denote the total number of memory records and dimension of each one. The \underline{me}mory latent representation $H^{(\text{me})}$ can be calculated by:
\begin{equation}
\left\{
\begin{aligned}
Q &= H^{(\text{sc})} \cdot W_Q + b_Q\\
\phi_j &= \frac{e^{Q \cdot \mathcal{M}[j]}}{\sum_{j=1}^m e^{Q \cdot \mathcal{M}[j]}}\\
V &= \sum_{j=1}^m \phi_j \cdot \mathcal{M}[j]\\
H^{(\text{me})} &= V \cdot W_V + b_V
\end{aligned}
\right.
\end{equation}
\noindent where $Q \in \mathbb{R}^{d_m}$ indicates the query vector projected from flattened $H^{(\text{sc})}$, $\phi_j$ is the attention score corresponding to $j^{\text{th}}$ memory record, and $V \in \mathbb{R}^{d_m}$ is the reconstructed prototype representation. Then, our HRA generates an \underline{aug}mented representation $H^{(\text{aug})}=[H^{(\text{sc})}, H^{(\text{me})}] \in \mathbb{R}^{L \times S \times (d_k+d_m)}$ by concatenating $H^{(\text{me})} \in \mathbb{R}^{L \times S \times d_m}$ with $H^{(\text{sc})}$.

\subsection{Predictor}
The final predictor consists of two fully connected layers with ReLU as an activation function, transforming the dimension of the hidden channel to the desired output dimension:
\begin{equation}
\left\{
\begin{aligned}
H^{(\text{out})} &= \text{relu}(H^{(\text{aug})}) \cdot W_{\text{o1}}+b_{\text{o1}}\\
\hat{Y} &= \text{relu}(H^{(\text{out})}) \cdot W_{\text{o2}} + b_{\text{o2}}
\end{aligned}
\right.
\end{equation}
\noindent where $W_{\text{o1}} \in \mathbb{R}^{(d_k+d_m) \times d_k}$, $W_{\text{o2}} \in \mathbb{R}^{d_k \times O}$, $b_{\text{o1}} \in \mathbb{R}^{d_k}$, and $b_{\text{o2}} \in \mathbb{R}^{O}$ are filter weights and biases. $\text{relu}(\cdot)$ denotes ReLU function that is applied element-wise. $\hat{Y} \in \mathbb{R}^{O \times L\times S}$ indicates the forecasting result of the next $O$ steps for $L$ locations and $S$ sources.

\noindent\textbf{Optimization.} In the learning phase, the Mean Squared Error (MSE) is used as the regression loss:
\begin{equation}
\mathcal{L}^{(\text{reg})} = \sum_{t=1}^{O} \sum_{l=1}^L \sum_{s=1}^S {\parallel Y_{t,l,s}-\hat{Y}_{t,l,s} \parallel}^2
\label{eq:mse}
\end{equation}
Eq.~\ref{eq:cluloss} and Eq.~\ref{eq:mse} are combined as the loss function for our TTS forecasting model:
\begin{equation}
\mathcal{L} = \mathcal{L}^{(\text{reg})} + \lambda \mathcal{L}^{(\text{cluster})}
\end{equation}

\noindent where $\lambda \ge 0$ is the balancing parameter.
\section{Experiments}
\begin{table}[h]
    \scriptsize
    \centering
    \setlength{\tabcolsep}{1mm}{
    \begin{tabular}{c||c|c}
        \hline
        \multicolumn{1}{c||}{Dataset}  & 
        \multicolumn{1}{c|}{\textbf{NYC Traffic Demand}} &
        \multicolumn{1}{c}{\textbf{BJ Air Quality}}
        \\
        \hline
        Time span &
        \multicolumn{1}{c|}{2015/10/24$\sim$2016/1/31} &
        \multicolumn{1}{c}{2013/3/1$\sim$2017/2/28}
        \\
        \hline
        Time slot &
        \multicolumn{1}{c|}{1 hour} &
        \multicolumn{1}{c}{1 hour} 
        \\
        \hline
        Timestep &
        \multicolumn{1}{c|}{4392} &
        \multicolumn{1}{c}{35064} 
        \\
        \hline
        Location &
        \multicolumn{1}{c|}{98 locations} &
        \multicolumn{1}{c}{10 locations}
        \\
        \hline
        Source & \{Bike, Taxi\}$\times$\{Inflow, Outflow\} & PM2.5, PM10, SO$_2$ 
        \\
        \hline
        \hline
        \# Train/Val/Test &
        \multicolumn{1}{c|}{3912/240/240} &
        \multicolumn{1}{c}{21048/7008/7008} 
        \\
        \hline
        \# Input length &
        \multicolumn{1}{c|}{16} &
        \multicolumn{1}{c}{16} 
        \\
        \hline
        \# Output length &
        \multicolumn{1}{c|}{1 /2 /3} &
        \multicolumn{1}{c}{1 /2 /3 }
        \\
        \hline
    \end{tabular}}
    \caption{Summary of Experimental Datasets}
    \label{tab:datasets}
\end{table}
\begin{table*}[t]
\scriptsize
\centering
\setlength{\tabcolsep}{0.5mm}{
\begin{tabular*}{17cm}{@{\extracolsep{\fill}}cc|ccc|ccc|ccc|ccc}
	\hline
	\multicolumn{1}{c|}{Source}  & 
	\multicolumn{2}{c|}{Bike Inflow} &
	\multicolumn{2}{c|}{Bike Outflow} &
	\multicolumn{2}{c|}{Taxi Inflow} & 
	\multicolumn{2}{c}{Taxi Outflow}
	\\
	\hline
	\multicolumn{1}{c|}{Method} &
	\multicolumn{1}{c|}{MAE} & 
	\multicolumn{1}{c|}{RMSE} &
	\multicolumn{1}{c|}{MAE} & 
	\multicolumn{1}{c|}{RMSE} &
	\multicolumn{1}{c|}{MAE} & 
	\multicolumn{1}{c|}{RMSE} &
	\multicolumn{1}{c|}{MAE} & 
	\multicolumn{1}{c}{RMSE}  
	\\
	\hline  
	\multicolumn{1}{c|}{LSTNet} &
	\multicolumn{1}{c|}{3.21 / 3.32 / 3.66} &
	\multicolumn{1}{c|}{8.65 / 9.05 / 9.72} &
	\multicolumn{1}{c|}{3.45 / 3.69 / 3.84} &
	\multicolumn{1}{c|}{9.40 / 10.14 / 10.43} &
	\multicolumn{1}{c|}{9.77 / 10.45 / 11.50} &
	\multicolumn{1}{c|}{20.66 / 21.75 / 23.81} &
	\multicolumn{1}{c|}{10.44 / 11.11 / 11.87} &
	\multicolumn{1}{c}{19.98 / 21.46 / 22.77}
	\\
	\multicolumn{1}{c|}{AGCRN} &
	\multicolumn{1}{c|}{2.40 / 2.70 / 3.02} &
	\multicolumn{1}{c|}{6.96 / 7.69 / 8.41} &
	\multicolumn{1}{c|}{2.60 / 2.97 / 3.31} &
	\multicolumn{1}{c|}{7.39 / 8.32 / 9.15} &
	\multicolumn{1}{c|}{\underline{7.08} / 8.10 / 9.16} &
	\multicolumn{1}{c|}{13.42 / \underline{15.71} / 18.16} &
	\multicolumn{1}{c|}{7.81 / 9.03 / 10.19} &
	\multicolumn{1}{c}{15.03 / \underline{17.41} / 19.29}
	\\
    \multicolumn{1}{c|}{MTGNN} &
	\multicolumn{1}{c|}{\underline{2.28} / \underline{2.44} / \underline{2.69}} &
	\multicolumn{1}{c|}{\underline{6.68} / \underline{7.05} / \underline{7.62}} &
	\multicolumn{1}{c|}{\underline{2.45} / \underline{2.60} / \underline{2.83}} &
	\multicolumn{1}{c|}{\underline{7.07} / \underline{7.46} / \underline{8.02}} &
	\multicolumn{1}{c|}{7.24 / \underline{8.04} / \underline{8.78}} &
	\multicolumn{1}{c|}{13.82 / 15.82 / \underline{17.71}} &
	\multicolumn{1}{c|}{7.97 / \underline{8.97} / \underline{9.96}} &
	\multicolumn{1}{c}{15.47 / 17.59 / \underline{19.26}}
	\\
	\multicolumn{1}{c|}{GWN} &
	\multicolumn{1}{c|}{2.37 / 2.66 / 2.91} &
	\multicolumn{1}{c|}{6.89 / 7.63 / 8.23} &
	\multicolumn{1}{c|}{2.57 / 2.91 / 3.17} &
	\multicolumn{1}{c|}{7.42 / 8.26 / 8.85} &
	\multicolumn{1}{c|}{7.00 / 8.23 / 9.59} &
	\multicolumn{1}{c|}{\underline{13.30} / 16.11 / 18.99} &
	\multicolumn{1}{c|}{\underline{7.72} / 9.05 / 10.29} &
	\multicolumn{1}{c}{\underline{14.98} / 17.48 / 19.69}
	\\
	\multicolumn{1}{c|}{StemGNN} &
	\multicolumn{1}{c|}{2.50 / 2.86 / 3.27} &
	\multicolumn{1}{c|}{7.17 / 8.04 / 9.01} &
	\multicolumn{1}{c|}{2.63 / 3.06 / 3.48} &
	\multicolumn{1}{c|}{7.55 / 8.63 / 9.65} &
	\multicolumn{1}{c|}{7.80 / 9.38 / 10.84} &
	\multicolumn{1}{c|}{14.76 / 18.01 / 21.12} &
	\multicolumn{1}{c|}{8.25 / 9.82 / 11.32} &
	\multicolumn{1}{c}{15.83 / 18.89 / 21.62}
	\\
	\multicolumn{1}{c|}{STtrans} &
	\multicolumn{1}{c|}{2.72 / 3.06 / 3.38} &
	\multicolumn{1}{c|}{7.74 / 8.54 / 9.35} &
	\multicolumn{1}{c|}{2.96 / 3.27 / 3.56} &
	\multicolumn{1}{c|}{8.33 / 8.93 / 9.67} &
	\multicolumn{1}{c|}{8.46 / 9.62 / 11.03} &
	\multicolumn{1}{c|}{15.67 / 17.74 / 20.42} &
	\multicolumn{1}{c|}{8.78 / 10.04 / 11.28} &
	\multicolumn{1}{c}{16.66 / 19.12 / 21.39}
	\\
	\multicolumn{1}{c|}{ST-Norm} &
	\multicolumn{1}{c|}{2.37 / 2.64 / 2.91} &
	\multicolumn{1}{c|}{6.93 / 7.54 / 8.13} &
	\multicolumn{1}{c|}{2.57 / 2.92 / 3.20} &
	\multicolumn{1}{c|}{7.42 / 8.29 / 8.95} &
	\multicolumn{1}{c|}{7.13 / 8.35 / 9.65} &
	\multicolumn{1}{c|}{13.60 / 16.28 / 18.72} &
	\multicolumn{1}{c|}{7.88 / 9.17 / 10.44} &
	\multicolumn{1}{c}{15.25 / 17.71 / 19.76}
	\\
	\multicolumn{1}{c|}{MiST} &
	\multicolumn{1}{c|}{2.32 / 2.52 / 2.76} &
	\multicolumn{1}{c|}{6.78 / 7.26 / 7.84} &
	\multicolumn{1}{c|}{2.57 / 2.83 / 3.01} &
	\multicolumn{1}{c|}{7.48 / 8.10 / 8.54} &
	\multicolumn{1}{c|}{7.93 / 8.73 / 9.61} &
	\multicolumn{1}{c|}{14.10 / 16.51 / 19.06} &
	\multicolumn{1}{c|}{8.44 / 9.19 / 10.82} &
	\multicolumn{1}{c}{15.56 / 17.93 / 19.33}
	\\
    \hline
	\multicolumn{1}{c|}{GMRL} &
	\multicolumn{1}{c|}{\textbf{2.03} / \textbf{2.09} / \textbf{2.17}} &
	\multicolumn{1}{c|}{\textbf{6.06} / \textbf{6.21} / \textbf{6.39}} &
	\multicolumn{1}{c|}{\textbf{2.14} / \textbf{2.22} / \textbf{2.32}} &
	\multicolumn{1}{c|}{\textbf{6.35} / \textbf{6.55} / \textbf{6.74}} &
	\multicolumn{1}{c|}{\textbf{6.49} / \textbf{7.14} / \textbf{7.56}} &
	\multicolumn{1}{c|}{\textbf{12.42} / \textbf{14.40} / \textbf{16.14}} &
	\multicolumn{1}{c|}{\textbf{7.07} / \textbf{7.79} / \textbf{8.34}} &
	\multicolumn{1}{c}{\textbf{13.94} / \textbf{15.26} / \textbf{16.61}}
	\\
	\hline
\end{tabular*}}
\caption{Performance on NYC Traffic Demand Dataset for the first/second/third horizon}
\label{tab:traf}
\end{table*}

\begin{table*}[t]
\scriptsize
\centering
\setlength{\tabcolsep}{0.5mm}{
\begin{tabular*}{15cm}{@{\extracolsep{\fill}}cc|ccc|ccc|ccc}
	\hline
	\multicolumn{1}{c|}{Source}  & 
	\multicolumn{2}{c|}{PM2.5} &
	\multicolumn{2}{c|}{PM10} &
	\multicolumn{2}{c}{SO$_2$} 
	\\
	\hline
	\multicolumn{1}{c|}{Method} &
	\multicolumn{1}{c|}{MAE} & 
	\multicolumn{1}{c|}{RMSE} &
	\multicolumn{1}{c|}{MAE} & 
	\multicolumn{1}{c|}{RMSE} &
	\multicolumn{1}{c|}{MAE} & 
	\multicolumn{1}{c}{RMSE}
	\\
	\hline  
	\multicolumn{1}{c|}{LSTNet} &
	\multicolumn{1}{c|}{12.98 / 17.55 / 21.76} &
	\multicolumn{1}{c|}{24.09 /  31.56 / 40.11} &
	\multicolumn{1}{c|}{19.23 / 24.93 / 30.26} &
	\multicolumn{1}{c|}{31.70 / 39.96 / 52.25} &
	\multicolumn{1}{c|}{2.15 /  3.69 / 4.47} &
	\multicolumn{1}{c}{7.13 /  10.07 / 11.41} 
	\\
	\multicolumn{1}{c|}{AGCRN} &
	\multicolumn{1}{c|}{9.85 / 15.39 / 19.83} &
	\multicolumn{1}{c|}{18.74 / 28.15 / 35.06} &
	\multicolumn{1}{c|}{15.62 / \underline{22.37} /        \textbf{27.32}} &
	\multicolumn{1}{c|}{26.41 / 36.13 / \underline{43.13}} &
	\multicolumn{1}{c|}{\underline{1.57} / 2.38 / 3.00} &
	\multicolumn{1}{c}{\underline{6.06} / 7.81 / 9.09} 
	\\
    \multicolumn{1}{c|}{MTGNN} &
	\multicolumn{1}{c|}{9.82 / 14.95 / \textbf{19.26}} &
	\multicolumn{1}{c|}{\underline{18.41} / \underline{27.35} / \underline{34.10}} &
	\multicolumn{1}{c|}{15.83 / 22.70 / 27.61} &
	\multicolumn{1}{c|}{26.30 / \underline{36.13} / 43.14} &
	\multicolumn{1}{c|}{1.73 / \underline{2.31} / \underline{2.82}} &
	\multicolumn{1}{c}{6.39 / \underline{7.68} / \underline{8.80}} 
	\\
	\multicolumn{1}{c|}{GWN} &
	\multicolumn{1}{c|}{\underline{9.78} / \underline{15.39} / 19.72} &
	\multicolumn{1}{c|}{18.57 / 28.08 / 35.10} &
	\multicolumn{1}{c|}{\underline{15.50} / 22.39 / 27.43} &
	\multicolumn{1}{c|}{\underline{26.09} / 36.22 / 43.76} &
	\multicolumn{1}{c|}{1.61 / 2.45 / 2.95} &
	\multicolumn{1}{c}{6.11 / 7.89 / 9.06} 
	\\
	\multicolumn{1}{c|}{StemGNN} &
	\multicolumn{1}{c|}{10.42 / 16.30 / 21.25} &
	\multicolumn{1}{c|}{19.53 / 29.30 / 36.71} &
	\multicolumn{1}{c|}{16.48 / 23.57 / 28.86} &
	\multicolumn{1}{c|}{27.71 / 37.84 / 45.29} &
	\multicolumn{1}{c|}{2.47 / 3.13 / 3.51} &
	\multicolumn{1}{c}{8.11 / 9.40 / 10.23} 
	\\
	\multicolumn{1}{c|}{STtrans} &
	\multicolumn{1}{c|}{10.55 / 15.46 / 19.56} &
	\multicolumn{1}{c|}{21.88 / 31.37 / 38.71} &
	\multicolumn{1}{c|}{16.61 / 23.14 / 28.21} &
	\multicolumn{1}{c|}{30.54 / 42.60 / 50.85} &
	\multicolumn{1}{c|}{2.34 / 3.13 / 3.79} &
	\multicolumn{1}{c}{7.55 / 9.41 / 10.71} 
	\\
	\multicolumn{1}{c|}{ST-Norm} &
	\multicolumn{1}{c|}{13.16 / 17.66 / 22.14} &
	\multicolumn{1}{c|}{21.42 / 29.98 / 36.89} &
	\multicolumn{1}{c|}{18.11 / 24.06 / 28.97} &
	\multicolumn{1}{c|}{28.02 / 37.57 / 44.79} &
	\multicolumn{1}{c|}{3.47 / 3.42 / 4.32} &
	\multicolumn{1}{c}{9.04 / 9.37 / 10.87} 
	\\
	\multicolumn{1}{c|}{MiST} &
	\multicolumn{1}{c|}{13.98 / 17.77 / 22.07} &
	\multicolumn{1}{c|}{21.66 / 30.77 / 35.95} &
	\multicolumn{1}{c|}{18.71 / 24.12 / 29.48} &
	\multicolumn{1}{c|}{28.20 / 38.44 / 46.42} &
	\multicolumn{1}{c|}{2.45 / 3.23 / 3.78} &
	\multicolumn{1}{c}{7.95  / 9.75 / 10.90} 
	\\
    \hline
	\multicolumn{1}{c|}{GMRL} &
	\multicolumn{1}{c|}{\textbf{9.53} / \textbf{14.80} / \underline{19.41}} &
	\multicolumn{1}{c|}{\textbf{18.29} / \textbf{26.89} / \textbf{33.82}} &
	\multicolumn{1}{c|}{\textbf{15.47} / \textbf{21.77} / \underline{27.37}} &
	\multicolumn{1}{c|}{\textbf{25.45} / \textbf{35.48} / \textbf{42.82}} &
	\multicolumn{1}{c|}{\textbf{1.55} / \textbf{2.28} / \textbf{2.77}} &
	\multicolumn{1}{c}{\textbf{5.88} / \textbf{6.61} / \textbf{8.19}} 
	\\
	\hline
\end{tabular*}}
\caption{Performance on BJ Air Quality Dataset for the first/second/third horizon}
\label{tab:air}
\end{table*}
\subsection{Experiment Setup}
\noindent \textbf{Datasets.} We conduct experiments on two real-world TTS datasets, namely NYC Traffic Demand and BJ Air Quality as listed in Table~\ref{tab:datasets}, the details of which are as follows:

\begin{itemize}
    \item \textbf{NYC Traffic Demand dataset}\footnote{\url{https://ride.citibikenyc.com/system-data}}\footnote{\url{https://www1.nyc.gov/site/tlc/about/tlc-trip-record-data.page}} is collected from NYC Bike Sharing System, which consists of 98 locations and four sources: Bike Inflow, Bike Outflow, Taxi Inflow, and Taxi Outflow. 
    \item \textbf{BJ Air Quality dataset}\footnote{\url{https://archive.ics.uci.edu/ml/datasets/Beijing+Multi-Site+Air-Quality+Data}} is collected from the Beijing Municipal Environmental Monitoring Center, which contains 10 locations and three pollutant sources: PM2.5, PM10, and SO$_2$.
\end{itemize}

\noindent \textbf{Settings.}
We implement the network with the Pytorch toolkit. For the model, the number of GMRE-TE layers and cluster components $K$ are set to 4 and 17. The kernel size of each dilated causal convolution component is 2, and the related expansion rate is \{2, 4, 8, 16\} in each GMRE-TE layer. This enables our model to handle the 16 input steps. The dimension of hidden channels $d_z$ is 24. The parameters for memory bank $m$ and $d_m$ are set to 8 and 48. The batch size is 8, and the learning rate of the Adam optimizer is 0.0001. In addition, the inputs are normalized by Z-Score.

\noindent \textbf{Baselines.} To quantitatively evaluate the prediction accuracy of our model, we implement eight baselines for comparison. 
\begin{itemize}
\item \textbf{LSTNet} \cite{lai2018modeling}. A deep neural network that combines convolutional neural networks and recurrent neural networks.
\item \textbf{MTGNN} \cite{wu2020connecting}. A spatial-temporal graph convolutional network that combines graph convolution and dilated convolution.
\item \textbf{Graph Wavenet} (GWN) \cite{wu2019graph}. A spatial-temporal network that combines adaptive graph convolutions with dilated casual convolution.
\item \textbf{AGCRN} \cite{bai2020adaptive}. A recurrent neural network with adaptive graph convolution. 
\item \textbf{StemGNN} \cite{cao2020spectral}.A spectral temporal graph neural network that maps temp-oral-spatial domain to spectral domain. 
\item \textbf{STtrans} \cite{wu2020hierarchically}. A deep neural network based on attention mechanism that integrates dynamic spatial, temporal and semantic dependencies.
\item \textbf{ST-Norm} \cite{deng2021st}. A normalization-based approach that refines the temporal components and the spatial components.
\item \textbf{MiST} \cite{huang2019mist}. A co-predictive model for multi-categorical abnormal events forecasting, that models the spatial and categorical embeddings incorporated.
\end{itemize}


\subsection{Overall Performance}
We evaluate the performance of our proposed model as well as the above baselines for tensor time series forecasting on all two datasets. We repeat the experiment five times for each model on each dataset and report the average results. Through Table~\ref{tab:traf}, we can find our model outperformed the state-of-the-arts to a large degree. For all four sources of the NYC Traffic Demand dataset, there have improvements of $\Delta$MAE\{14.88\%, 15.10\%, 11.14\%, 12.61\%\}, and $\Delta$RMSE\{12.45\%, 12.78\%, 7.94\%, 11.02\%\} on three horizons average. Among the co-prediction models, MiST performed better than STtrans. However, MiST is second only to MTGNN in terms of bike demand and much inferior in taxi demand. Because it requires reference sources as auxiliary input while outputting the target source prediction, it is not adequate to quantify heterogeneity on three variables (i.e., the time, the location, and the source variable). By contrast, StemGNN and ST-Norm gave a worse performance because ignoring the role of source variables too much makes them not applicable for tensor time series forecasting. 

The comparison results with baselines on the BJ Air Quality dataset are shown in Table~\ref{tab:air}. GMRL has the improvements of $\Delta$MAE\{1.87\%, 0.90\%, 1.45\%\}, and $\Delta$RMSE\{1.05\%, 1.66\%, 7.94\%\} for three sources on three horizons average. The BJ Air Quality dataset only has 10 locations and less corrections, and many weather-related features are unavailable. This causes GMRL to perform less well on data with insufficient information, but it fully exploits the correlations among available features. In this case, baseline models are still not as efficient as GMRL.

\subsection{Ablation Study}
To make a thorough evaluation of key components in our model, we create several variants as follows: 
\begin{itemize}
    \item \textbf{w/o GMRE.} It removes GMRE in each GMRE-TE layer and keeps other components of GMRL, i.e., TTSE is directly used as input to Temporal Encoder.
    \item \textbf{w/ KMeans.} It applies the KMeans clustering algorithm instead of GMRE to obtain a cluster-wise representation, which randomly initializes K cluster centers and calculates the categories according to the distance between the samples and the centers.
    \item \textbf{w/o HRA.} It removes HRA in GMRL, i.e., $H^\text{(aug)}=H^\text{(sc)}$, which means the representation $H^\text{(sc)}$ is directly fed into Predictor to generate the predictions.
    \item \textbf{w/o $\mathcal{L}^{(\text{cluster})}$.} It removes $\mathcal{L}^{(\text{cluster})}$ in GMRE (Eq.~\ref{eq:cluloss}), i.e., $\mathcal{L} = \mathcal{L}^{(\text{reg})}$ in the learning phase.
\end{itemize}
\begin{table}[t]
\scriptsize
\centering
\setlength{\tabcolsep}{1mm}{
\begin{tabular}{cc|ccc|ccc|ccc}
	\hline
	\multicolumn{1}{c|}{Source}  & 
	\multicolumn{1}{c|}{Bike Inflow} &
	\multicolumn{1}{c|}{Bike Outflow} &
	\multicolumn{1}{c|}{Taxi Inflow} & 
	\multicolumn{1}{c}{Taxi Outflow}
	\\
	\hline
	\multicolumn{1}{c|}{Variant} &
	\multicolumn{1}{c|}{MAE / RMSE} & 
        \multicolumn{1}{c|}{MAE / RMSE} & 
        \multicolumn{1}{c|}{MAE / RMSE} & 
        \multicolumn{1}{c}{MAE / RMSE} 
	\\
	\hline  
        \multicolumn{1}{c|}{w/o GMRE} &
	\multicolumn{1}{c|}{2.65 / 7.25} &
        \multicolumn{1}{c|}{2.71 / 7.45} &
        \multicolumn{1}{c|}{9.16 / 18.00} &
        \multicolumn{1}{c}{9.40 / 17.87}
	\\
         \multicolumn{1}{c|}{w/ KMeans} &
	\multicolumn{1}{c|}{3.23 / 8.84} &
        \multicolumn{1}{c|}{3.27 / 8.93} &
        \multicolumn{1}{c|}{10.63 / 20.48} &
        \multicolumn{1}{c}{10.82 / 20.42}
	\\
        \multicolumn{1}{c|}{w/o HRA} &
	\multicolumn{1}{c|}{2.42 / 6.71} &
        \multicolumn{1}{c|}{2.55 / 7.06} &
        \multicolumn{1}{c|}{7.95 / 16.51} &
        \multicolumn{1}{c}{8.81 / 16.95}
	\\
        \multicolumn{1}{c|}{w/o $\mathcal{L}^{(\text{cluster})}$} &
	\multicolumn{1}{c|}{2.28 / 6.64} &
        \multicolumn{1}{c|}{2.41 / 7.02} &
        \multicolumn{1}{c|}{7.93 / 16.37} &
        \multicolumn{1}{c}{8.54 / 16.86}
	\\
	\multicolumn{1}{c|}{GMRL} &
	\multicolumn{1}{c|}{\textbf{2.17} / \textbf{6.39}} &
        \multicolumn{1}{c|}{\textbf{2.32} / \textbf{6.74}} &
        \multicolumn{1}{c|}{\textbf{7.56} / \textbf{16.14}} &
        \multicolumn{1}{c}{\textbf{8.34} / \textbf{16.61}}
	\\
	\hline
\end{tabular}}
\caption{Ablation Study}
\label{tab:ablation}
\end{table}
Through Table~\ref{tab:ablation}, we can see that compared to w/o GMRE, w/ KMeans will deteriorate the forecasting performance, because it only assembles instances by hard clustering, and the clustering result is affected by the selection of the initial cluster center. Removing HRA and $\mathcal{L}^{(\text{cluster})}$ deteriorate performance to a lesser degree. But still, there is a significant gap compared with the full GMRL. All these demonstrate that the proposed GMRL is a complete and indivisible set. 
\subsection{Case Study}
\begin{figure}[t]
  \centering
  \includegraphics[width=\linewidth]{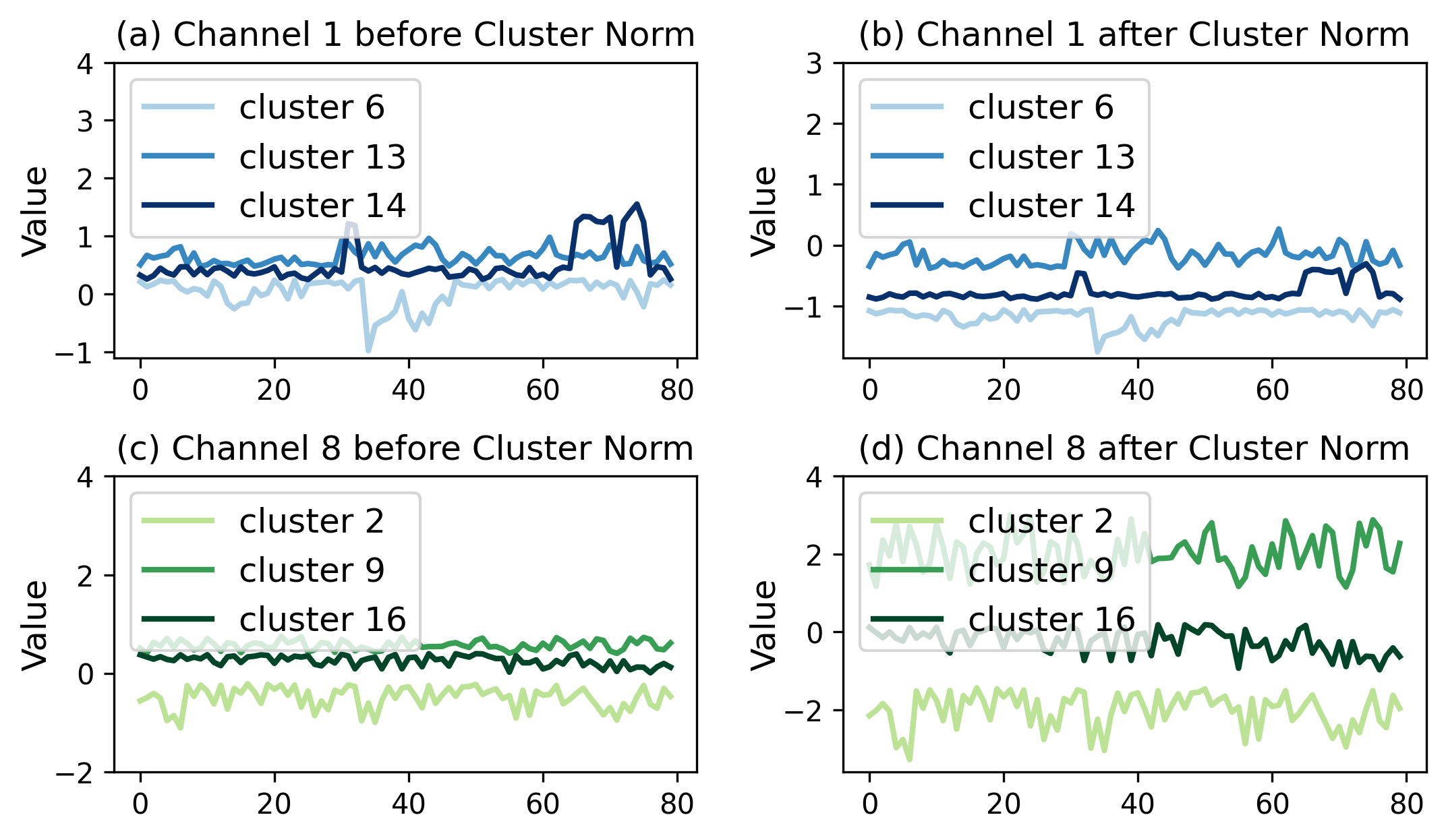}
  \caption{Case studies on cluster variations in different channels before and after Cluster Norm.}
  \label{fig:casestudy}
\end{figure}
\subsubsection{Hidden Representations before/after Cluster Norm}
In this subsection, we conduct multiple studies to qualitatively analyze the representations learned from GMRE of the last GMRE-TE layer. In Figure~\ref{fig:casestudy}, we visualize variations of clusters in different channels before and after applying Cluster Norm. In Figure~\ref{fig:casestudy}a, the evolution curves of clusters 2, 13, and 14 in Channel 1 overlap and intertwine with each other, especially for clusters 13 and 14. This is because clusters 13 and 14 both originate from the same source (Bike Inflow), and the overlapping features correspond to several identical locations. As shown in Figure~\ref{fig:casestudy}b, after Cluster Norm, the evolution curves of clusters in Channel 1 become distinguishable. The clusters corresponding to spatial identifiers maintain a stable behavior over time, which remains unchanged even after Cluster Norm. The performance of clusters in Channel 8 before and after Cluster Norm is similar to that in Channel 1, as shown in Figure~\ref{fig:casestudy}c and Figure~\ref{fig:casestudy}d. However, in Channel 8, clusters 2, 9, and 16 represent temporal identifiers, and they exhibit changes with varying frequencies over time compared to Channel 1. Clearly, after applying Cluster Norm, each cluster in different channels exhibits heterogeneous evolution with sufficient discriminability. This study confirms the component clustering capability and time-adaptability of GMRL.
\subsubsection{Gaussian Mixture Parameters}
\begin{figure}[t]
  \centering
  \includegraphics[width=\linewidth]{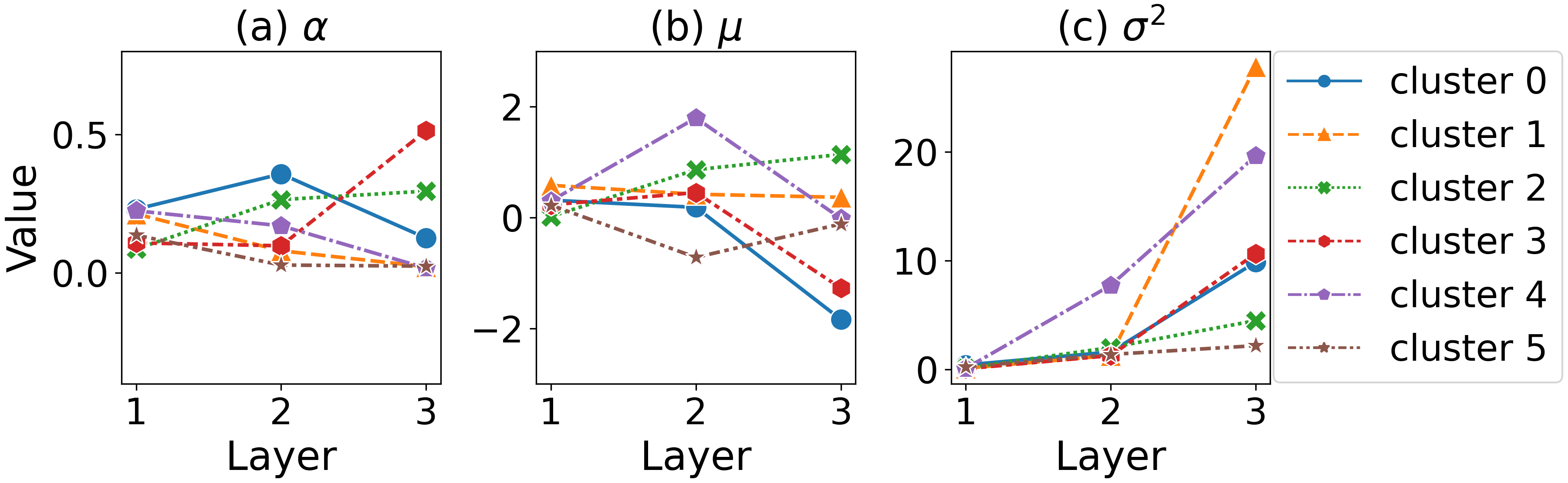}
  \caption{Case studies on the Gaussian mixture parameters $\alpha$, $\mu$, and $\sigma^2$ in each GMRE-TE layer.}
  \label{fig:para}
\end{figure}

In Figure~\ref{fig:para}, we display multiple Gaussian mixture parameters learned in GMRE. As the GMRE-TE layer goes deeper, Gaussian mixture parameters $\alpha$, $\mu$, and $\sigma^2$ can perfectly describe six heterogeneous components after passing through the last GMRE-TE layer. For example, cluster components 4 (purple) and 5 (brown) tend to be similar on parameters $\alpha$ and $\mu$, but their variances are very different. When the models converge, the Gaussian distribution they determine is sufficient to represent heterogeneous components.
\subsection{Hyper-parameter Study}
\begin{figure}[t]
  \centering
  \includegraphics[width=\linewidth]{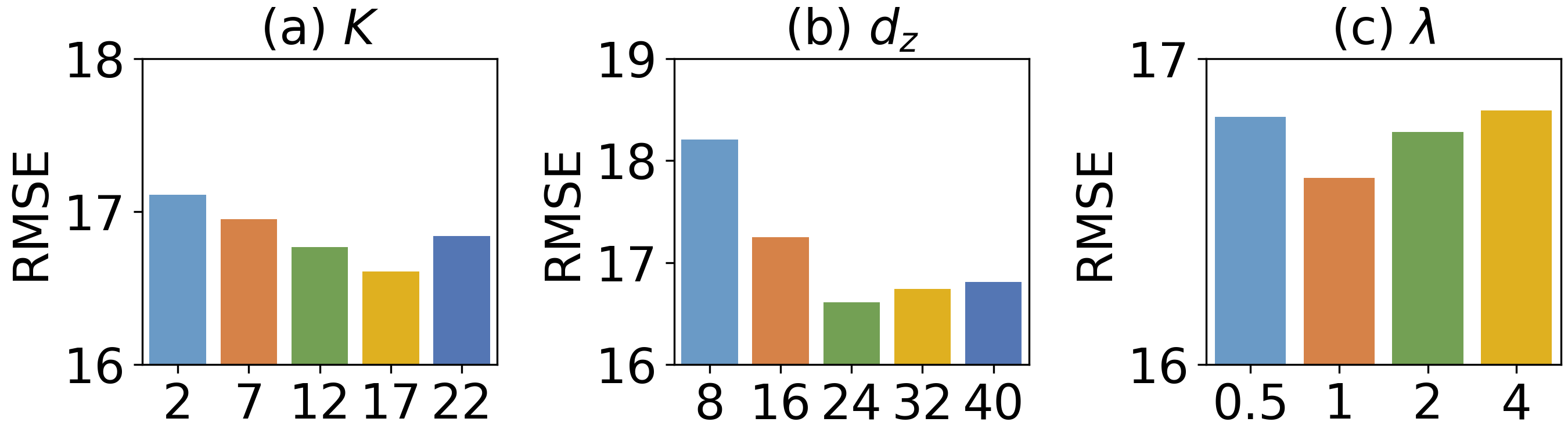}
  \caption{Hyper-parameter studies for the number of cluster components $K$, the dimension of hidden channels $d_z$, and the balancing parameter $\lambda$.}
  \label{fig:hyper}
\end{figure}

We further study the effect of key hyper-parameters manually set in the proposed GMRL, including the number of cluster components $K$, the dimension of the hidden channel $d_z$, and the balance parameter $\lambda$, as shown in Figure~\ref{fig:hyper}. We find that as $K$ and $\lambda$ increase, the prediction error of GMRL first decreases and then increases. This phenomenon follows our understanding that the model can suffer from underfitting or overfitting due to improper settings of cluster components and balance parameter. Surprisingly, the dimension of the hidden channel has little effect on GMRL: setting it to 24 can achieve fairly good performance, and increasing the dimension of the hidden channel hardly changes the performance.
\section{Related Work}
Time series forecasting has been studied for decades and applied to various fields. Some studies utilized the attention mechanism for dealing with lost memories \cite{li2019enhancing,vaswani2017attention,zhou2021informer}. To extract complex spatial-temporal patterns, existing works applied various operations over temporal and spatial domains. Specifically, the graph neural network is developed to model the spatial relation \cite{yu2017spatio,li2017diffusion,oreshkin2021fc,shang2021discrete,chen2020multi}; the attention mechanism and its variants contain spatial attention \cite{fang2019gstnet,zheng2020gman} and temporal attention \cite{wu2021autoformer,zheng2020gman,liu2021pyraformer,zhou2022fedformer}; the convolution operator is leveraged to spatial convolution \cite{deng2022graph,guo2021hierarchical}, temporal convolution \cite{wu2020connecting,wu2019graph}, spatial-temporal convolution \cite{guo2019deep,yang2021space} and adaptive convolution \cite{pan2019urban}. \cite{zhu2021mixseq} proposes an end2end mixture model to cluster microscopic series for macroscopic time series forecasting. \cite{jiang2021dl} reviews the deep learning models and builds a standard benchmark. \cite{deng2021st,deng2022multi} refine the high-frequency and local components from MTS data by using normalization technology. Unlike MTS involved data within a matrix, data points within TTS usually take on more substantial heterogeneity and dynamics. Since such unique properties of TTS compared to MTS, traditional methods for predicting MTS (or MTS with reference source) are not entirely suitable for TTS prediction. The inability to explicitly learn and encode interactions from TTS brings up some intuitive concerns. A new line of research further extends the spatial-temporal dependency modeling by considering another domain (i.e., source or categorical domain) \cite{ye2019co,huang2019mist,huang2021cross,han2021dynamic,wang2021spatio} or extra data source \cite{jiang2023learning}. Our work distinguishes itself from these methods in the following aspects: (i) our GMRE is dynamic, which could explicitly represent the complicated heterogeneous components among multiple variables in TTS; (ii) our HRA enhances the learned representations with sequence-specific global patterns to distinguish and generalize to diverse scenarios, which could help model adaptability to TTS.
\section{Conclusion}
In this paper, we develop a dynamic Gaussian mixture representations learning framework GMRL, which seeks to individually model each heterogeneity component implied in multiple variables. In addition, we design an HRA module that enhances the learned representation with sequence-specific global patterns to distinguish and generalize to diverse scenarios. Extensive experiments on two real-world datasets verify our perspective's insightfulness and the proposed approach's effectiveness. In the next step, we plan to improve the speed of adaptation and efficiency of GMRL. 
\section*{Contribution Statement}
Renhe Jiang is the corresponding author for this study.

\bibliographystyle{named}
\bibliography{ijcai23}

\begin{thebibliography}{}

\bibitem[\protect\citeauthoryear{Bai \bgroup \em et al.\egroup
  }{2020}]{bai2020adaptive}
Lei Bai, Lina Yao, Can Li, Xianzhi Wang, and Can Wang.
\newblock Adaptive graph convolutional recurrent network for traffic
  forecasting.
\newblock {\em Advances in neural information processing systems},
  33:17804--17815, 2020.

\bibitem[\protect\citeauthoryear{Cao \bgroup \em et al.\egroup
  }{2020}]{cao2020spectral}
Defu Cao, Yujing Wang, Juanyong Duan, Ce~Zhang, Xia Zhu, Congrui Huang, Yunhai
  Tong, Bixiong Xu, Jing Bai, Jie Tong, et~al.
\newblock Spectral temporal graph neural network for multivariate time-series
  forecasting.
\newblock {\em Advances in neural information processing systems},
  33:17766--17778, 2020.

\bibitem[\protect\citeauthoryear{Chen \bgroup \em et al.\egroup
  }{2020}]{chen2020multi}
Weiqi Chen, Ling Chen, Yu~Xie, Wei Cao, Yusong Gao, and Xiaojie Feng.
\newblock Multi-range attentive bicomponent graph convolutional network for
  traffic forecasting.
\newblock In {\em Proceedings of the AAAI conference on artificial
  intelligence}, volume~34, pages 3529--3536, 2020.

\bibitem[\protect\citeauthoryear{Deng \bgroup \em et al.\egroup
  }{2021}]{deng2021st}
Jinliang Deng, Xiusi Chen, Renhe Jiang, Xuan Song, and Ivor~W Tsang.
\newblock St-norm: Spatial and temporal normalization for multi-variate time
  series forecasting.
\newblock In {\em Proceedings of the 27th ACM SIGKDD conference on knowledge
  discovery \& data mining}, pages 269--278, 2021.

\bibitem[\protect\citeauthoryear{Deng \bgroup \em et al.\egroup
  }{2022a}]{deng2022multi}
Jinliang Deng, Xiusi Chen, Renhe Jiang, Xuan Song, and Ivor~W Tsang.
\newblock A multi-view multi-task learning framework for multi-variate time
  series forecasting.
\newblock {\em IEEE Transactions on Knowledge and Data Engineering}, 2022.

\bibitem[\protect\citeauthoryear{Deng \bgroup \em et al.\egroup
  }{2022b}]{deng2022graph}
Leyan Deng, Defu Lian, Zhenya Huang, and Enhong Chen.
\newblock Graph convolutional adversarial networks for spatiotemporal anomaly
  detection.
\newblock {\em IEEE Transactions on Neural Networks and Learning Systems},
  33(6):2416--2428, 2022.

\bibitem[\protect\citeauthoryear{Duan \bgroup \em et al.\egroup
  }{2022}]{duan2022plae}
Jufang Duan, Yi~Wang, and Wei Zheng.
\newblock Plae: Time-series prediction improvement by adaptive decomposition.
\newblock In {\em Pacific Rim International Conference on Artificial
  Intelligence}, pages 394--407. Springer, 2022.

\bibitem[\protect\citeauthoryear{Fang \bgroup \em et al.\egroup
  }{2019}]{fang2019gstnet}
Shen Fang, Qi~Zhang, Gaofeng Meng, Shiming Xiang, and Chunhong Pan.
\newblock Gstnet: Global spatial-temporal network for traffic flow prediction.
\newblock In {\em IJCAI}, pages 2286--2293, 2019.

\bibitem[\protect\citeauthoryear{Fang \bgroup \em et al.\egroup
  }{2021}]{fang2021mdtp}
Ziquan Fang, Lu~Pan, Lu~Chen, Yuntao Du, and Yunjun Gao.
\newblock Mdtp: a multi-source deep traffic prediction framework over
  spatio-temporal trajectory data.
\newblock {\em Proceedings of the VLDB Endowment}, 14(8):1289--1297, 2021.

\bibitem[\protect\citeauthoryear{Guo \bgroup \em et al.\egroup
  }{2019}]{guo2019deep}
Shengnan Guo, Youfang Lin, Shijie Li, Zhaoming Chen, and Huaiyu Wan.
\newblock Deep spatial--temporal 3d convolutional neural networks for traffic
  data forecasting.
\newblock {\em IEEE Transactions on Intelligent Transportation Systems},
  20(10):3913--3926, 2019.

\bibitem[\protect\citeauthoryear{Guo \bgroup \em et al.\egroup
  }{2021}]{guo2021hierarchical}
Kan Guo, Yongli Hu, Yanfeng Sun, Sean Qian, Junbin Gao, and Baocai Yin.
\newblock Hierarchical graph convolution network for traffic forecasting.
\newblock In {\em Proceedings of the AAAI Conference on Artificial
  Intelligence}, volume~35, pages 151--159, 2021.

\bibitem[\protect\citeauthoryear{Han \bgroup \em et al.\egroup
  }{2021}]{han2021dynamic}
Liangzhe Han, Bowen Du, Leilei Sun, Yanjie Fu, Yisheng Lv, and Hui Xiong.
\newblock Dynamic and multi-faceted spatio-temporal deep learning for traffic
  speed forecasting.
\newblock In {\em Proceedings of the 27th ACM SIGKDD Conference on Knowledge
  Discovery \& Data Mining}, pages 547--555, 2021.

\bibitem[\protect\citeauthoryear{Huang \bgroup \em et al.\egroup
  }{2019}]{huang2019mist}
Chao Huang, Chuxu Zhang, Jiashu Zhao, Xian Wu, Dawei Yin, and Nitesh Chawla.
\newblock Mist: A multiview and multimodal spatial-temporal learning framework
  for citywide abnormal event forecasting.
\newblock In {\em The World Wide Web Conference}, pages 717--728, 2019.

\bibitem[\protect\citeauthoryear{Huang \bgroup \em et al.\egroup
  }{2021}]{huang2021cross}
Chao Huang, Chuxu Zhang, Peng Dai, and Liefeng Bo.
\newblock Cross-interaction hierarchical attention networks for urban anomaly
  prediction.
\newblock In {\em Proceedings of the Twenty-Ninth International Conference on
  International Joint Conferences on Artificial Intelligence}, pages
  4359--4365, 2021.

\bibitem[\protect\citeauthoryear{Jiang \bgroup \em et al.\egroup
  }{2021}]{jiang2021dl}
Renhe Jiang, Du~Yin, Zhaonan Wang, Yizhuo Wang, Jiewen Deng, Hangchen Liu,
  Zekun Cai, Jinliang Deng, Xuan Song, and Ryosuke Shibasaki.
\newblock Dl-traff: Survey and benchmark of deep learning models for urban
  traffic prediction.
\newblock In {\em Proceedings of the 30th ACM international conference on
  information \& knowledge management}, pages 4515--4525, 2021.

\bibitem[\protect\citeauthoryear{Jiang \bgroup \em et al.\egroup
  }{2022}]{jiang2022spatio}
Renhe Jiang, Zhaonan Wang, Jiawei Yong, Puneet Jeph, Quanjun Chen, Yasumasa
  Kobayashi, Xuan Song, Shintaro Fukushima, and Toyotaro Suzumura.
\newblock Spatio-temporal meta-graph learning for traffic forecasting.
\newblock {\em arXiv preprint arXiv:2211.14701}, 2022.

\bibitem[\protect\citeauthoryear{Jiang \bgroup \em et al.\egroup
  }{2023}]{jiang2023learning}
Renhe Jiang, Zhaonan Wang, Yudong Tao, Chuang Yang, Xuan Song, Ryosuke
  Shibasaki, Shu-Ching Chen, and Mei-Ling Shyu.
\newblock Learning social meta-knowledge for nowcasting human mobility in
  disaster.
\newblock In {\em Proceedings of the ACM Web Conference 2023}, pages
  2655--2665, 2023.

\bibitem[\protect\citeauthoryear{Jing \bgroup \em et al.\egroup
  }{2021}]{jing2021network}
Baoyu Jing, Hanghang Tong, and Yada Zhu.
\newblock Network of tensor time series.
\newblock In {\em Proceedings of the Web Conference 2021}, pages 2425--2437,
  2021.

\bibitem[\protect\citeauthoryear{Lai \bgroup \em et al.\egroup
  }{2018}]{lai2018modeling}
Guokun Lai, Wei-Cheng Chang, Yiming Yang, and Hanxiao Liu.
\newblock Modeling long-and short-term temporal patterns with deep neural
  networks.
\newblock In {\em The 41st international ACM SIGIR conference on research \&
  development in information retrieval}, pages 95--104, 2018.

\bibitem[\protect\citeauthoryear{Li \bgroup \em et al.\egroup
  }{2017}]{li2017diffusion}
Yaguang Li, Rose Yu, Cyrus Shahabi, and Yan Liu.
\newblock Diffusion convolutional recurrent neural network: Data-driven traffic
  forecasting.
\newblock {\em arXiv preprint arXiv:1707.01926}, 2017.

\bibitem[\protect\citeauthoryear{Li \bgroup \em et al.\egroup
  }{2019}]{li2019enhancing}
Shiyang Li, Xiaoyong Jin, Yao Xuan, Xiyou Zhou, Wenhu Chen, Yu-Xiang Wang, and
  Xifeng Yan.
\newblock Enhancing the locality and breaking the memory bottleneck of
  transformer on time series forecasting.
\newblock {\em Advances in neural information processing systems}, 32, 2019.

\bibitem[\protect\citeauthoryear{Liu \bgroup \em et al.\egroup
  }{2021}]{liu2021pyraformer}
Shizhan Liu, Hang Yu, Cong Liao, Jianguo Li, Weiyao Lin, Alex~X Liu, and
  Schahram Dustdar.
\newblock Pyraformer: Low-complexity pyramidal attention for long-range time
  series modeling and forecasting.
\newblock In {\em International conference on learning representations}, 2021.

\bibitem[\protect\citeauthoryear{Oreshkin \bgroup \em et al.\egroup
  }{2021}]{oreshkin2021fc}
Boris~N Oreshkin, Arezou Amini, Lucy Coyle, and Mark Coates.
\newblock Fc-gaga: Fully connected gated graph architecture for spatio-temporal
  traffic forecasting.
\newblock In {\em Proceedings of the AAAI Conference on Artificial
  Intelligence}, volume~35, pages 9233--9241, 2021.

\bibitem[\protect\citeauthoryear{Pan \bgroup \em et al.\egroup
  }{2019}]{pan2019urban}
Zheyi Pan, Yuxuan Liang, Weifeng Wang, Yong Yu, Yu~Zheng, and Junbo Zhang.
\newblock Urban traffic prediction from spatio-temporal data using deep meta
  learning.
\newblock In {\em Proceedings of the 25th ACM SIGKDD international conference
  on knowledge discovery \& data mining}, pages 1720--1730, 2019.

\bibitem[\protect\citeauthoryear{Rogers \bgroup \em et al.\egroup
  }{2013}]{rogers2013multilinear}
Mark Rogers, Lei Li, and Stuart~J Russell.
\newblock Multilinear dynamical systems for tensor time series.
\newblock {\em Advances in Neural Information Processing Systems}, 26, 2013.

\bibitem[\protect\citeauthoryear{Shang \bgroup \em et al.\egroup
  }{2021}]{shang2021discrete}
Chao Shang, Jie Chen, and Jinbo Bi.
\newblock Discrete graph structure learning for forecasting multiple time
  series.
\newblock {\em arXiv preprint arXiv:2101.06861}, 2021.

\bibitem[\protect\citeauthoryear{Shao \bgroup \em et al.\egroup
  }{2022}]{shao2022decoupled}
Zezhi Shao, Zhao Zhang, Wei Wei, Fei Wang, Yongjun Xu, Xin Cao, and Christian~S
  Jensen.
\newblock Decoupled dynamic spatial-temporal graph neural network for traffic
  forecasting, 2022.

\bibitem[\protect\citeauthoryear{van~den Oord \bgroup \em et al.\egroup
  }{2016}]{Oord2016WaveNetAG}
A{\"a}ron van~den Oord, Sander Dieleman, Heiga Zen, Karen Simonyan, Oriol
  Vinyals, Alex Graves, Nal Kalchbrenner, Andrew~W. Senior, and Koray
  Kavukcuoglu.
\newblock Wavenet: A generative model for raw audio.
\newblock In {\em SSW}, 2016.

\bibitem[\protect\citeauthoryear{Vaswani \bgroup \em et al.\egroup
  }{2017}]{vaswani2017attention}
Ashish Vaswani, Noam Shazeer, Niki Parmar, Jakob Uszkoreit, Llion Jones,
  Aidan~N Gomez, {\L}ukasz Kaiser, and Illia Polosukhin.
\newblock Attention is all you need.
\newblock {\em Advances in neural information processing systems}, 30, 2017.

\bibitem[\protect\citeauthoryear{Wang \bgroup \em et al.\egroup
  }{2021}]{wang2021spatio}
Zhaonan Wang, Renhe Jiang, Zekun Cai, Zipei Fan, Xin Liu, Kyoung-Sook Kim, Xuan
  Song, and Ryosuke Shibasaki.
\newblock Spatio-temporal-categorical graph neural networks for fine-grained
  multi-incident co-prediction.
\newblock In {\em Proceedings of the 30th ACM international conference on
  information \& knowledge management}, pages 2060--2069, 2021.

\bibitem[\protect\citeauthoryear{Wang \bgroup \em et al.\egroup
  }{2022}]{wang2022event}
Zhaonan Wang, Renhe Jiang, Hao Xue, Flora~D Salim, Xuan Song, and Ryosuke
  Shibasaki.
\newblock Event-aware multimodal mobility nowcasting.
\newblock In {\em Proceedings of the AAAI Conference on Artificial
  Intelligence}, volume~36, pages 4228--4236, 2022.

\bibitem[\protect\citeauthoryear{Wu \bgroup \em et al.\egroup
  }{2019}]{wu2019graph}
Zonghan Wu, Shirui Pan, Guodong Long, Jing Jiang, and Chengqi Zhang.
\newblock Graph wavenet for deep spatial-temporal graph modeling.
\newblock {\em arXiv preprint arXiv:1906.00121}, 2019.

\bibitem[\protect\citeauthoryear{Wu \bgroup \em et al.\egroup
  }{2020a}]{wu2020hierarchically}
Xian Wu, Chao Huang, Chuxu Zhang, and Nitesh~V Chawla.
\newblock Hierarchically structured transformer networks for fine-grained
  spatial event forecasting.
\newblock In {\em Proceedings of The Web Conference 2020}, pages 2320--2330,
  2020.

\bibitem[\protect\citeauthoryear{Wu \bgroup \em et al.\egroup
  }{2020b}]{wu2020connecting}
Zonghan Wu, Shirui Pan, Guodong Long, Jing Jiang, Xiaojun Chang, and Chengqi
  Zhang.
\newblock Connecting the dots: Multivariate time series forecasting with graph
  neural networks.
\newblock In {\em Proceedings of the 26th ACM SIGKDD international conference
  on knowledge discovery \& data mining}, pages 753--763, 2020.

\bibitem[\protect\citeauthoryear{Wu \bgroup \em et al.\egroup
  }{2021}]{wu2021autoformer}
Haixu Wu, Jiehui Xu, Jianmin Wang, and Mingsheng Long.
\newblock Autoformer: Decomposition transformers with auto-correlation for
  long-term series forecasting.
\newblock {\em Advances in Neural Information Processing Systems},
  34:22419--22430, 2021.

\bibitem[\protect\citeauthoryear{Yang \bgroup \em et al.\egroup
  }{2021}]{yang2021space}
Song Yang, Jiamou Liu, and Kaiqi Zhao.
\newblock Space meets time: Local spacetime neural network for traffic flow
  forecasting.
\newblock In {\em 2021 IEEE International Conference on Data Mining (ICDM)},
  pages 817--826. IEEE, 2021.

\bibitem[\protect\citeauthoryear{Ye \bgroup \em et al.\egroup
  }{2019}]{ye2019co}
Junchen Ye, Leilei Sun, Bowen Du, Yanjie Fu, Xinran Tong, and Hui Xiong.
\newblock Co-prediction of multiple transportation demands based on deep
  spatio-temporal neural network.
\newblock In {\em Proceedings of the 25th ACM SIGKDD International Conference
  on Knowledge Discovery \& Data Mining}, pages 305--313, 2019.

\bibitem[\protect\citeauthoryear{Yu \bgroup \em et al.\egroup
  }{2017}]{yu2017spatio}
Bing Yu, Haoteng Yin, and Zhanxing Zhu.
\newblock Spatio-temporal graph convolutional networks: A deep learning
  framework for traffic forecasting.
\newblock {\em arXiv preprint arXiv:1709.04875}, 2017.

\bibitem[\protect\citeauthoryear{Zhang \bgroup \em et al.\egroup
  }{2016}]{zhang2016dnn}
Junbo Zhang, Yu~Zheng, Dekang Qi, Ruiyuan Li, and Xiuwen Yi.
\newblock Dnn-based prediction model for spatio-temporal data.
\newblock In {\em Proceedings of the 24th ACM SIGSPATIAL international
  conference on advances in geographic information systems}, pages 1--4, 2016.

\bibitem[\protect\citeauthoryear{Zhang \bgroup \em et al.\egroup
  }{2017}]{zhang2017deep}
Junbo Zhang, Yu~Zheng, and Dekang Qi.
\newblock Deep spatio-temporal residual networks for citywide crowd flows
  prediction.
\newblock In {\em Thirty-first AAAI conference on artificial intelligence},
  2017.

\bibitem[\protect\citeauthoryear{Zhao \bgroup \em et al.\egroup
  }{2018}]{zhao2018unsupervised}
Tiancheng Zhao, Kyusong Lee, and Maxine Eskenazi.
\newblock Unsupervised discrete sentence representation learning for
  interpretable neural dialog generation.
\newblock {\em arXiv preprint arXiv:1804.08069}, 2018.

\bibitem[\protect\citeauthoryear{Zheng \bgroup \em et al.\egroup
  }{2020}]{zheng2020gman}
Chuanpan Zheng, Xiaoliang Fan, Cheng Wang, and Jianzhong Qi.
\newblock Gman: A graph multi-attention network for traffic prediction.
\newblock In {\em Proceedings of the AAAI conference on artificial
  intelligence}, volume~34, pages 1234--1241, 2020.

\bibitem[\protect\citeauthoryear{Zhou \bgroup \em et al.\egroup
  }{2021}]{zhou2021informer}
Haoyi Zhou, Shanghang Zhang, Jieqi Peng, Shuai Zhang, Jianxin Li, Hui Xiong,
  and Wancai Zhang.
\newblock Informer: Beyond efficient transformer for long sequence time-series
  forecasting.
\newblock In {\em Proceedings of the AAAI Conference on Artificial
  Intelligence}, volume~35, pages 11106--11115, 2021.

\bibitem[\protect\citeauthoryear{Zhou \bgroup \em et al.\egroup
  }{2022}]{zhou2022fedformer}
Tian Zhou, Ziqing Ma, Qingsong Wen, Xue Wang, Liang Sun, and Rong Jin.
\newblock Fedformer: Frequency enhanced decomposed transformer for long-term
  series forecasting.
\newblock In {\em International Conference on Machine Learning}, pages
  27268--27286. PMLR, 2022.

\bibitem[\protect\citeauthoryear{Zhu \bgroup \em et al.\egroup
  }{2021}]{zhu2021mixseq}
Zhibo Zhu, Ziqi Liu, Ge~Jin, Zhiqiang Zhang, Lei Chen, Jun Zhou, and Jianyong
  Zhou.
\newblock Mixseq: Connecting macroscopic time series forecasting with
  microscopic time series data.
\newblock {\em Advances in Neural Information Processing Systems},
  34:12904--12916, 2021.

\end{thebibliography}

\end{document}